\documentclass{llncs}
\usepackage[dvips]{graphicx}
\usepackage{epsfig}
\usepackage{ae}
\usepackage{url}
\begin{document}

\pagestyle{empty}
\mainmatter
\title{On the Success Rate of Crossover Operators for Genetic Programming with Offspring Selection}
	\author{
		Gabriel Kronberger \and Stephan Winkler\and Michael Affenzeller \and \\ Andreas Beham \and Stefan Wagner
	}
	\institute{
		Heuristic and Evolutionary Algorithms Laboratory\\
		School of Informatics, Communications and Media - Hagenberg\\
		Upper Austria University of Applied Sciences\\
		Softwarepark 11, A-4232 Hagenberg, Austria\\
		\email{\{gkronber,swinkler,maffenze,abeham,swagner\}@heuristiclab.com}
	}

\maketitle

\renewcommand{\thefootnote}{} \footnotetext{\hspace{-0em}The final
  publication is available at \url{http://link.springer.com/chapter/10.1007/978-3-642-04772-5_102}}

\renewcommand\thefootnote{\arabic{footnote}}

%%%%%%%%%%%%%%%%%%%%%%%%%%%%%%%%%%%%%%%%%%%%%%%%%%%%%%%%%%%%%%%%%%%%%%%%%%%%%%%%
%%
%% Abstract
%%
%%%%%%%%%%%%%%%%%%%%%%%%%%%%%%%%%%%%%%%%%%%%%%%%%%%%%%%%%%%%%%%%%%%%%%%%%%%%%%%%
\begin{abstract}
Genetic programming is a powerful heuristic search technique that is
used for a number of real world applications to solve amongst others
regression, classification, and time-series forecasting problems. A
lot of progress towards a theoretic description of genetic programming
in form of schema theorems has been made, but the internal dynamics
and success factors of genetic programming are still not fully
understood. In particular, the effects of different crossover
operators in combination with offspring selection are largely unknown.

This contribution sheds light on the ability of well-known GP
crossover operators to create better offspring when applied to
benchmark problems. We conclude that standard (sub-tree swapping)
crossover is a good default choice in combination with offspring
selection, and that GP with offspring selection and random selection
of crossover operators can improve the performance of the algorithm in
terms of best solution quality when no solution size constraints are
applied.
\end{abstract}

%%%%%%%%%%%%%%%%%%%%%%%%%%%%%%%%%%%%%%%%%%%%%%%%%%%%%%%%%%%%%%%%%%%%%%%%%%%%%%%%
%%
%% Genetic Programming
%%
%%%%%%%%%%%%%%%%%%%%%%%%%%%%%%%%%%%%%%%%%%%%%%%%%%%%%%%%%%%%%%%%%%%%%%%%%%%%%%%%
\section{Genetic Programming}
Genetic programming (GP) is a generalization of genetic algorithms
first studied at length by John Koza \cite{koza1992}. Whereas the goal
of genetic algorithms is to find a fixed length vector of symbols that
encodes a solution to the problem, the goal of genetic programming is
to find a variable-length program that solves the original problem
when executed. Common practice is to use a tree-based representation
of computer programs similar to so called symbolic expressions of
functional programming languages such as LISP.

Genetic programming is a powerful heuristic search method that has
been used successfully to solve real world problems from various
application domains, including classification, regression, and
forecasting of time-series \cite{langdon2002,poli2008}.

Offspring selection \cite{affenzeller2005a} is a generic selection
concept for evolutionary algorithms that aims to reduce the effect of
premature convergence often observed with traditional selection
operators by preservation of important alleles
\cite{affenzeller2009}. The main difference to the usual definition of
evolutionary algorithms is that after parent selection, recombination
and optional mutation, offspring selection filters the newly generated
solutions. Only solutions that have a better quality than their best
parent are added to the next generation of the population. In this
aspect offspring selection is similar to non-destructive crossover
\cite{Soule1997}, soft brood selection \cite{Altenberg1994}, and
hill-climbing crossover \cite{oreilly1995}. Non-destructive crossover
compares the quality of one child to the quality of the parent and
adds the better one to the next generation, whereas offspring
selection generates new children until a successful offspring is
found. Soft brood selection generates $n$ offspring and uses
tournament selection to determine the individual that is added to the
next generation, but in comparison to offspring selection the children
do not compete against the parents. Hill-climbing crossover generates
new offspring from the parents as long as better solutions can be
found. The best solution found by this hill-climbing scheme is added
to the next generation. The recently described hereditary selection
concept \cite{murphy2008gecco,murphy2008} also uses a similar
offspring selection scheme in combination with parent selection that
is biased to select solutions with few common ancestors.

%%%%%%%%%%%%%%%%%%%%%%%%%%%%%%%%%%%%%%%%%%%%%%%%%%%%%%%%%%%%%%%%%%%%%%%%%%%%%%%%
%%
%% Motivation
%%
%%%%%%%%%%%%%%%%%%%%%%%%%%%%%%%%%%%%%%%%%%%%%%%%%%%%%%%%%%%%%%%%%%%%%%%%%%%%%%%%
\section{Motivation}
Since the very first experiments with genetic programming a lot of
effort has been put into the definition of a theoretic foundation for
GP in order to gain a better understanding of its internal dynamics. A
lot of progress \cite{langdon2002,poli2003a,poli2003,poli2002} towards
the definition of schema theorems for variable length genetic
programming and sub-tree swapping crossover, as well as homologous
crossover operators \cite{poli2004} has been made. Still, an overall
understanding of the internal dynamics and the success factors of
genetic programming is still missing. The effects of mixed or variable
arity function sets or different mutation operators in combination
with more advanced selection schemes are still not fully
understood. In particular, the effects of different crossover
operators on the tree size and solution quality in combination with
offspring selection are largely unknown.

In this research we aim to shed light on the effects of GP crossover
operators regarding their ability to create improved solutions in the
context of offspring selection. We apply GP with offspring selection
to three benchmark problems: symbolic regression (Poly-10), time
series prediction (Mackey-Glass) and classification (Wisconsin
diagnostic breast cancer). The same set of experiments was also
executed for the 4-bit even parity problem, but because of space
constraints the results of those experiments are not reported in this
paper.

Recently we have analyzed the success rate of GP crossover operators
with offspring selection with strict solution size constraints
\cite{kronberger2009}. In the paper at hand we report results of
similar experiments with the same set of crossover operators and
benchmark problems, but without strict solution size constraints.

%%%%%%%%%%%%%%%%%%%%%%%%%%%%%%%%%%%%%%%%%%%%%%%%%%%%%%%%%%%%%%%%%%%%%%%%%%%%%%%%
%%
%% Configuration of Experiments
%%
%%%%%%%%%%%%%%%%%%%%%%%%%%%%%%%%%%%%%%%%%%%%%%%%%%%%%%%%%%%%%%%%%%%%%%%%%%%%%%%%
\section{Configuration of Experiments}
The crossover operators used in the experiments are: standard
(sub-tree swapping) \cite{koza1992} \cite{poli2002}, one-point
\cite{langdon2002}, uniform \cite{poli1998}, size-fair, homologous,
and size-fair \cite{langdon2000fairxo}. Additionally, the same
experiments were also executed with a crossover variant that chooses
one of the five crossover operators randomly for each crossover event
\cite{kronberger2009}.  Except for the crossover operator, the problem
specific evaluation operator, and the function set all other
parameters of the algorithm were the same for all experiments. The
random initial population was generated with probabilistic tree
creation (PTC2) \cite{luke2000} and uniform distribution of tree sizes
in the interval $[3;50]$. A single-point mutation operator was used to
manipulate 15\% of the solution candidates by exchanging either a
function symbol (50\%) or a terminal symbol (50\%). See
Table~\ref{tab:problem-params} for a summary of all GP parameters.

To analyze the results, the quality of the best solution, average tree
size in the whole population as well as offspring selection pressure
were logged at each generation step together with the number of
solutions that have been evaluated so far. Each run was stopped as
soon as the maximal offspring selection pressure or the maximal number
of solution evaluations was reached.

Offspring selection pressure of a population is defined as the ratio
of the number of solution evaluations that were necessary to fill the
population to the population size \cite{affenzeller2005a}. High
offspring selection pressure means that the chance that crossover
generates better children is very small, whereas low offspring
selection pressure means that the crossover operator can easily
generate better children.

% Poly-10
\subsection{Symbolic Regression -- Poly-10}
The Poly-10 symbolic regression benchmark problem uses ten input
variables $x_1,\dots,x_{10}$. The function for the target variable $y$
is defined as $y = x_1x_2 + x_3x_4 + x_5x_6 + x_1x_7x_9 +
x_3x_6x_{10}$ \cite{langdon2005,poli2003b}. For our experiments 100
training samples were generated randomly by sampling the values for
the input variables uniformly in the range $[-1,1[$. The usual
    function set of +,-,*, \% (protected division) and the terminal
    set of $x_1\dots,x_{10}$ without constants was used.  The mean
    squared errors function (MSE) over all 100 training samples was
    used as fitness function.

% Mackey-Glass
\subsection{Time Series Prediction -- Mackey-Glass}
The Mackey-Glass ($\tau = 17$)\footnote{Data set available from:
  http://neural.cs.nthu.edu.tw/jang/benchmark/} chaotic time series is
an artificial benchmark data set sometimes used as a representative
time series for medical or financial data sets \cite{langdon2005}. We
used the first 928 samples as training set, the terminal set for the
prediction of $x(t)$ consisted of past observations
$x_{128},x_{64},x_{32},x_{16},x_{8},x_{4},x_{2},x_1$ and integer
constants in the interval $[1;127]$. The function set and the fitness
function (MSE) were the same as in the experiments for Poly-10.

% Wisconsin
\subsection{Classification -- Wisconsin Diagnostic Breast Cancer}
The Wisconsin diagnostic breast cancer data set from the UCI Machine
Learning Repository \cite{uci2007} is a well known data set for binary
classification. Only a part (400 samples) of the whole data set was
used and the values of the target variable were transformed to values
2 and 4. Before each genetic programming run the whole data set was
shuffled, thus the training set was different for each run.

Again the mean squared errors function for the whole training set was
used as fitness function. In contrast to the previous experiments a
rather large function set was used that included functions with
different arities and types (see Table~\ref{tab:problem-params}). The
terminal set consisted of all ten input variables and real-valued
constants in the interval $[-20;20]$.

\begin{table}
  \begin{center}
    \scriptsize
    \begin{tabular}{|l|l|l|}
      \hline
      General parameters & Population size & 1000 \\
      for all experiments & Initialization & PTC2 (uniform $[3..50]$) \\
      & Parent selection & fitness-proportional (50\%), random (50\%)\\
      & & strict offspring selection, 1-elitism\\
      & Mutation rate & 15\% single point (50\% functions, 50\% terminals)\\
      & constraints & unlimited tree size and depth \\
      \hline
      \hline
      Poly-10 & Function set & ADD, SUB, MUL, DIV (protected) \\
      & Terminal set & $x_1 \dots x_{10}$ \\
      & Fitness function & Mean squared errors \\
      & Max. evaluations & 1.000.000 \\
      \hline
      \hline
      Mackey-Glass & Function set & ADD, SUB, MUL, DIV (protected) \\
      & Terminal set & $x_{128}, x_{64}, \dots, x_2, x_1$, constants: $1..127$\\
      & Fitness function & Mean squared errors \\
      & Max. evaluations & 5.000.000 \\
      \hline
      \hline
      Wisconsin & Function set & ADD, MUL, SUB, DIV (protected),\\
      & & LOG, EXP, SIGNUM, SIN, COS, TAN,\\
      & & IF-THEN-ELSE, LESS-THAN, GREATER-THAN,\\
      & & EQUAL, NOT, AND, OR, XOR \\
      & Terminal set & $x_{1}, \dots, x_{10}$, constants: $[-20..20]$\\
      & Fitness function & Mean squared errors \\
      & Max. evaluations & 2.000.000 \\
      \hline
    \end{tabular}
    \normalsize
    \caption{General parameters for all experiments and specific parameters for each benchmark problem.}
    \label{tab:problem-params}
  \end{center}
\end{table}

%%%%%%%%%%%%%%%%%%%%%%%%%%%%%%%%%%%%%%%%%%%%%%%%%%%%%%%%%%%%%%%%%%%%%%%%%%%%%%%%
%%
%% Results
%%
%%%%%%%%%%%%%%%%%%%%%%%%%%%%%%%%%%%%%%%%%%%%%%%%%%%%%%%%%%%%%%%%%%%%%%%%%%%%%%%%
\section{Results}
\begin{figure}
  \begin{center}
    \scriptsize
    \begin{tabular}{p{10mm}cccccc}
      & Standard & Onepoint & Uniform & Size-fair & Homologous & Mixed\\
      & \resizebox{16mm}{!}{\includegraphics{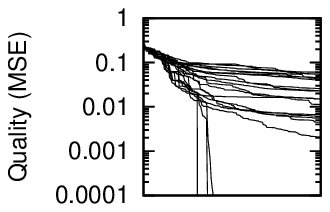}} &
      \resizebox{16mm}{!}{\includegraphics{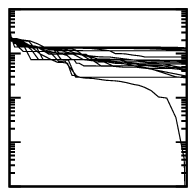}} &
      \resizebox{16mm}{!}{\includegraphics{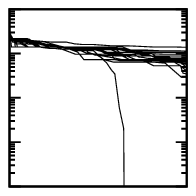}} &
      \resizebox{16mm}{!}{\includegraphics{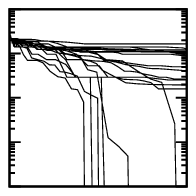}} &
      \resizebox{16mm}{!}{\includegraphics{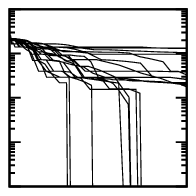}} &
      \resizebox{16mm}{!}{\includegraphics{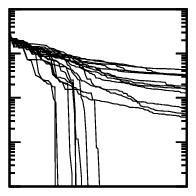}} \\ \\
      & \resizebox{16mm}{!}{\includegraphics{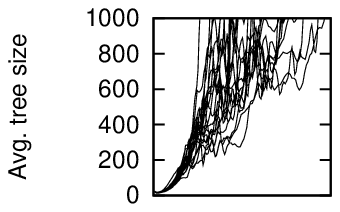}} &
      \resizebox{16mm}{!}{\includegraphics{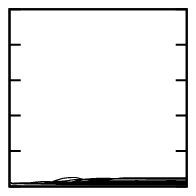}} &
      \resizebox{16mm}{!}{\includegraphics{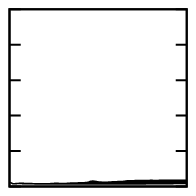}} &
      \resizebox{16mm}{!}{\includegraphics{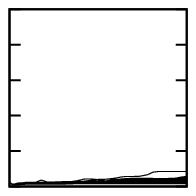}} &
      \resizebox{16mm}{!}{\includegraphics{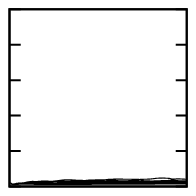}} &
      \resizebox{16mm}{!}{\includegraphics{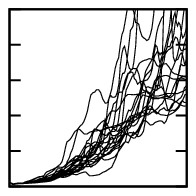}} \\ \\
      & \resizebox{16mm}{!}{\includegraphics{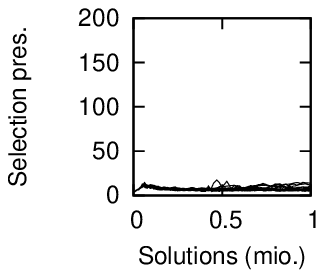}} &
      \resizebox{16mm}{!}{\includegraphics{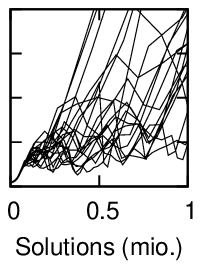}} &
      \resizebox{16mm}{!}{\includegraphics{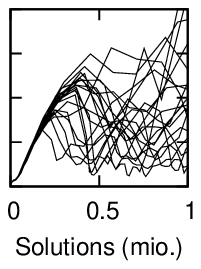}} &
      \resizebox{16mm}{!}{\includegraphics{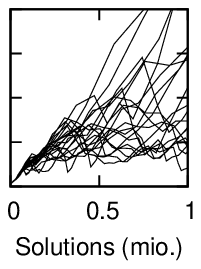}} &
      \resizebox{16mm}{!}{\includegraphics{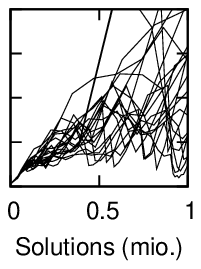}} &
      \resizebox{16mm}{!}{\includegraphics{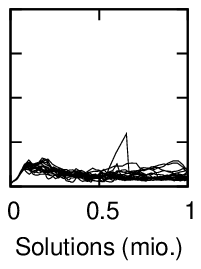}} \\ \\
    \end{tabular}
    \footnotesize
    \caption{Best solution quality (MSE, note log scale), average tree
      size, and offspring selection pressure for 20 runs with each
      crossover operator for the Poly-10 problem.}
    \label{fig:poly-10-single}
  \end{center}
\end{figure}

Figure~\ref{fig:poly-10-single} shows the quality progress (MSE, note
log scale), average tree size, and offspring selection pressure for
each of the six crossover operators over time (number of evaluated
solutions). The first row shows the best solution quality, the second
row shows average tree size over the whole population and the third
row shows offspring selection pressure.

Size-fair, homologous, and mixed crossover are the most successful
operators, whereas onepoint and uniform crossover show rather bad
performance. The average tree size grows exponentially in the
experiments with standard and mixed crossover, whereas with onepoint,
uniform, size-fair and homologous crossover the average tree size
stays at a low level. The most interesting result is that offspring
selection pressure stays at a low level over the whole run when
standard or mixed crossover are used. Offspring selection pressure
rises gradually over the whole run when standard crossover is used
with size constraints \cite{kronberger2009}. The different behavior
when no size constraints are applied indicates that larger offspring
solutions are more likely to be better than their parent solutions
than offspring solutions of equal or smaller size. The offspring
selection pressure charts for onepoint, uniform, size-fair and
homologous crossover show the usual effect, namely that it becomes
increasingly harder for crossover to produce successful children.

\begin{figure}
  \scriptsize
  \begin{center}
    \begin{tabular}{p{10mm}cccccc}
      & Standard & Onepoint & Uniform & Size-fair & Homologous & Mixed \\
      & \resizebox{16mm}{!}{\includegraphics{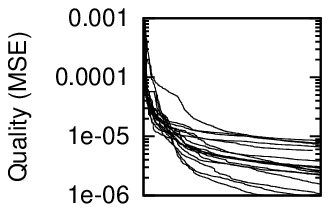}} &
      \resizebox{16mm}{!}{\includegraphics{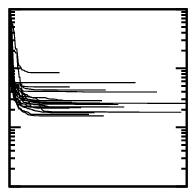}} &
      \resizebox{16mm}{!}{\includegraphics{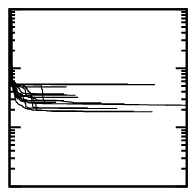}} &
      \resizebox{16mm}{!}{\includegraphics{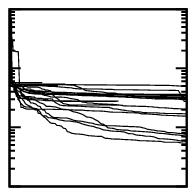}} &
      \resizebox{16mm}{!}{\includegraphics{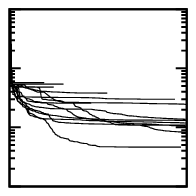}} &
      \resizebox{16mm}{!}{\includegraphics{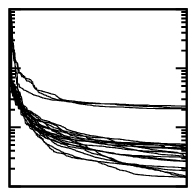}} \\ \\
      & \resizebox{16mm}{!}{\includegraphics{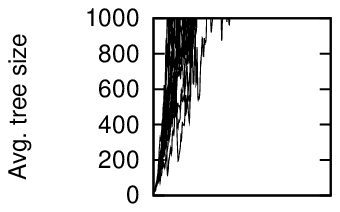}} &
      \resizebox{16mm}{!}{\includegraphics{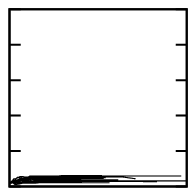}} &
      \resizebox{16mm}{!}{\includegraphics{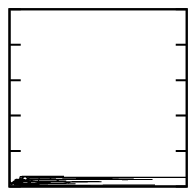}} &
      \resizebox{16mm}{!}{\includegraphics{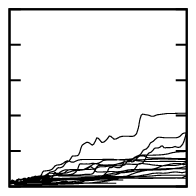}} &
      \resizebox{16mm}{!}{\includegraphics{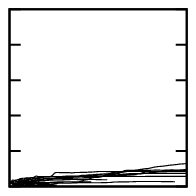}} &
      \resizebox{16mm}{!}{\includegraphics{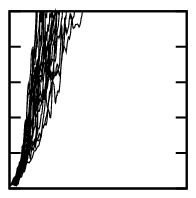}} \\ \\
      & \resizebox{16mm}{!}{\includegraphics{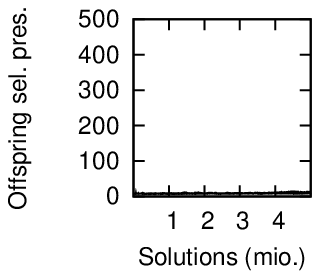}} &
      \resizebox{16mm}{!}{\includegraphics{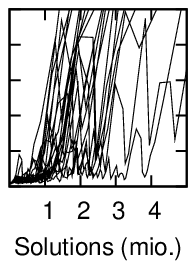}} &
      \resizebox{16mm}{!}{\includegraphics{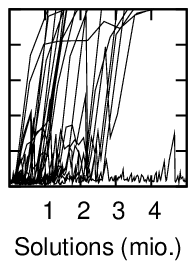}} &
      \resizebox{16mm}{!}{\includegraphics{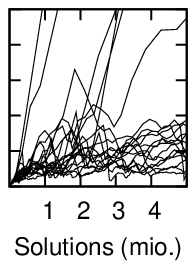}} &
      \resizebox{16mm}{!}{\includegraphics{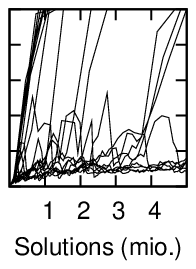}} &
      \resizebox{16mm}{!}{\includegraphics{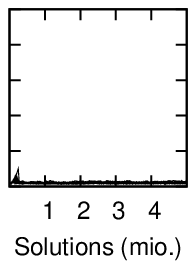}} \\ \\
    \end{tabular}
    \footnotesize
    \caption{Best solution quality (MSE, note log scale), average tree
      size, and offspring selection pressure (note different scale)
      for 20 runs with each crossover operator for the Mackey-Glass
      problem.}
    \label{fig:mackey-glass-single}
  \end{center}
\end{figure}

Figure~\ref{fig:mackey-glass-single} shows the results for the
Mackey-Glass problem. Standard crossover and mixed crossover show good
performance in terms of solution quality and the expected exponential
growth of solution size. Size-fair crossover had similar behavior as
homologous crossover. Onepoint and uniform crossover are the least
effective operators. The offspring selection pressure charts show that
with onepoint and uniform crossover the offspring selection pressure
rises quickly. The runs with standard crossover and mixed crossover
again have low offspring selection pressure over the whole run.

\begin{figure}
  \scriptsize
  \begin{center}
    \begin{tabular}{p{10mm} cccccc}
      & Standard & Onepoint & Uniform & Size-fair & Homologous & Mixed \\
      & \resizebox{16mm}{!}{\includegraphics{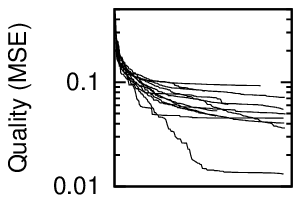}} &
      \resizebox{16mm}{!}{\includegraphics{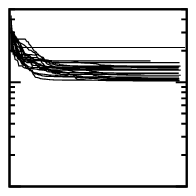}} &
      \resizebox{16mm}{!}{\includegraphics{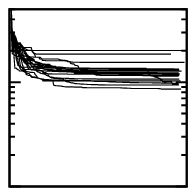}} &
      \resizebox{16mm}{!}{\includegraphics{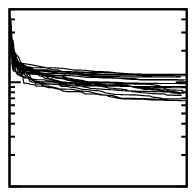}} &
      \resizebox{16mm}{!}{\includegraphics{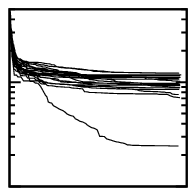}} &
      \resizebox{16mm}{!}{\includegraphics{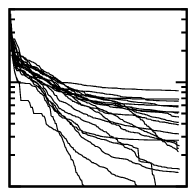}} \\ \\
      & \resizebox{16mm}{!}{\includegraphics{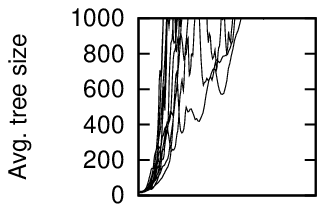}} &
      \resizebox{16mm}{!}{\includegraphics{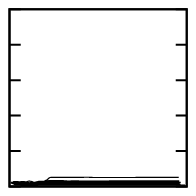}} &
      \resizebox{16mm}{!}{\includegraphics{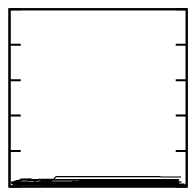}} &
      \resizebox{16mm}{!}{\includegraphics{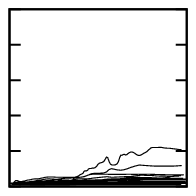}} &
      \resizebox{16mm}{!}{\includegraphics{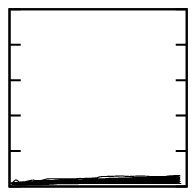}} &
      \resizebox{16mm}{!}{\includegraphics{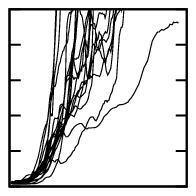}} \\ \\
      & \resizebox{16mm}{!}{\includegraphics{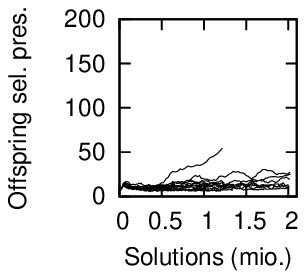}} &
      \resizebox{16mm}{!}{\includegraphics{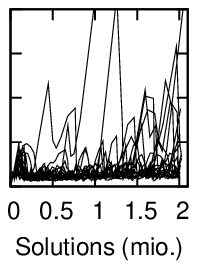}} &
      \resizebox{16mm}{!}{\includegraphics{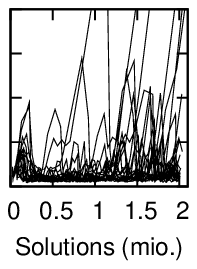}} &
      \resizebox{16mm}{!}{\includegraphics{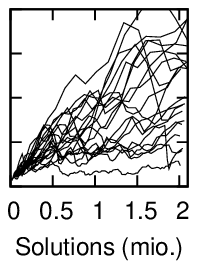}} &
      \resizebox{16mm}{!}{\includegraphics{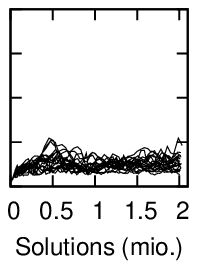}} &
      \resizebox{16mm}{!}{\includegraphics{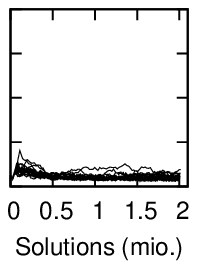}} \\ \\
    \end{tabular}
    \footnotesize
    \caption{Best solution quality (MSE, note log scale), average tree
      size, and offspring selection pressure for 20 runs with each
      crossover operator for the Wisconsin classification problem.}
    \label{fig:wisconsin-single}
  \end{center}
\end{figure}

Figure~\ref{fig:wisconsin-single} shows the results for the Wisconsin
classification problem. Mixed crossover performs better than standard
crossover for this problem. Onepoint, uniform, size-fair, and
homologous crossover reached similar solution quality, except for one
outlier with homologous crossover. The offspring selection pressure
curves of onepoint and uniform crossover show that offspring selection
pressure remains at a low level until a point of convergence is
reached where the offspring selection pressure rapidly increases to
the upper limit. The explanation for this is that onepoint and uniform
crossover cause convergence to a fixed tree shape. When all solutions
have the same tree shape it becomes very hard to find better
solutions. Only the runs with size-fair crossover show the usual
pattern of gradually increasing offspring selection pressure.  An
interesting result is that offspring selection pressure also remains
low for homologous crossover even though it doesn't show the
exponential growth in solution size as standard and mixed
crossover. The flat offspring selection pressure curve could be caused
by either the extended function set or the structure of the data
set. Further investigations are necessary to fully explain this
observation.

\section{Conclusion}
Based on the results for the benchmark problems it can be concluded
that standard (sub-tree swapping) crossover is a good default
choice. The results also show that onepoint and uniform crossover
operators do not perform very well on their own. They also have the
tendency to quickly freeze the tree shape, and should be combined with
mutation operators which manipulate tree shape.
 
The aim of the experiments with the mixed-crossover variant was to
find out if a combination of all five crossover operators in one GP
run has a beneficial effect either in terms of achievable solution
quality or efficiency. For two of the three benchmark problems the
runs with mixed crossover found better solutions than runs with
standard crossover. This result is in contrast to the results of
experiments with strict size constraints where runs with mixed
crossover did not find better solutions than runs with standard
crossover \cite{kronberger2009}.

\section{Acknowledgment}
This work mainly reflects research work done within the Josef
Ressel-center for heuristic optimization ``Heureka!'' at the Upper
Austria University of Applied Sciences, Campus Hagenberg. The center
``Heureka!'' is supported by the Austrian Research Promotion Agency
(FFG) on behalf of the Austrian Federal Ministry of Economy, Family
and Youth (BMWFJ). G.K. thanks the participants of EuroGP 2009 for the
insightful comments and discussions.  \bibliographystyle{plain}
\bibliography{Literature}
\end{document}